\documentclass[envcountsame,runningheads]{llncs}

\usepackage{a4,a4wide}

\usepackage{graphicx}
\usepackage{amssymb}
\usepackage{epstopdf}
\usepackage{tikz}
\usepackage{amsmath}
\usepackage{wasysym} % for ligthening
\usepackage{listings}
\usepackage{color}
\usepackage{rotating}
\usepackage{appendix}
\usepackage{url}

\usepackage{algorithmic}
\usepackage{algorithm}
\usepackage{caption}
\usepackage{hyperref}
\usepackage{xspace}

\usepackage{wrapfig}

\algsetup{indent=1ex} 

\usepackage{tikz}
\usetikzlibrary{positioning}

\newcommand{\RISS}{\textit{riss}\xspace}

\def\mynode#{\vtop \bgroup \hsize 0pt \parindent 0pt 
        \rightskip = 0pt minus \maxdimen \let\next=}

\title{Coprocessor - a Standalone SAT Preprocessor}
\author{Norbert Manthey}
\titlerunning{{}}
\authorrunning{{}}
\institute{Knowledge Representation and Reasoning Group\\
Technische Universit\"at Dresden, 01062 Dresden, Germany\\
\email{norbert@janeway.inf.tu-dresden.de}}
\begin{document}
\maketitle

\pagestyle{empty}

\begin{abstract}
%TODO: suit for improvements, results and comparison method!
In this work a stand-alone preprocessor for SAT is presented that is able to perform most of the known preprocessing techniques. Preprocessing a formula in SAT is important for performance since redundancy can be removed. The preprocessor is part of the SAT solver \RISS~\cite{riss-2011} and is called \textit{Coprocessor}. Not only \RISS, but also \textit{MiniSat 2.2}~\cite{minisat22} benefit from it, because the \textit{SatELite} preprocessor of \textit{MiniSat} does not implement recent techniques. By using more advanced techniques, \textit{Coprocessor} is able to reduce the redundancy in a formula further and improves the overall solving performance.% Thus, \textit{Coprocessor} can replace it as the next standard SAT preprocessor.
\end{abstract}

\vspace{-1.5ex}
\section{Introduction}
In theory SAT problems with $n$ variables have a worst case execution time of $O(2^{n})$~\cite{cook71-complexity}. Reducing the number of variables results in a theoretically faster search. However, in practice the number of variables does not correlate with the runtime. The number of clauses highly influences the performance of unit propagation. 
%Reducing the number of variables by increasing the number of clauses usually does not boost the performance of the solver.
Preprocessing helps to reduce the size of the formula by removing variables and clauses that are redundant. Due to limited space it is assumed that the reader is familiar with basic preprocessing techniques~\cite{satelite}. Preprocessing techniques can be classified into two categories: Techniques, which change a formula in a way that the satisfying assignment for the preprocessed formula is not necessarily a model for the original formula, are called \textit{satisfiability-preserving techniques}. Thus, for these techniques undo information has to be stored. For the second category, this information is not required. The second category is called \textit{equivalence-preserving techniques}, because the preprocessed and original formula are equivalent.

This paper is structured in the following way. An overview of the implemented techniques is given in Sect.~\ref{techniques}. Details on \textit{Coprocessor}, a format for storing the undo information and a comparison to \textit{SatELite} is given in Sect.~\ref{coprocessor}. Finally, a conclusion is given in Sect.~\ref{conclusion}.

\vspace{-1.5ex}
\section{Preprocessor Techniques} \label{techniques}
The notation used to describe the preprocessor is the following: variables are numbers and literals are positive or negative variables, e.g. $2$ and $\neg2$. A clause $C$ is a disjunction of a set of literals, denoted by $[l_1, \dots, l_n]$. A formula is a conjunction of clauses. The original formula will be referred to as $F$, the preprocessed formula is always called $F'$. Unit propagation on $F$ is denoted by BCP($l$), where $l$ is the literal that is assigned to \textit{true}.

%The following techniques are integrated in \textit{Coprocessor} and are described briefly below: Variable Elimination, Blocked Clause Elimination, Equivalence Elimination, Hidden Tautology Elimination, Failed Literal Probing, Clause Vivification, Extended Resolution and Unit Propagation.

%\vspace{-1ex}
\subsection{Satisfiability-Preserving Techniques}

The following techniques change $F$ in a way, that models of $F'$ are no model for $F$ anymore. Therefore, these methods need to store undo information. Undoing of these methods has to be done carefully, because the order influences the resulting assignment. All the elimination steps have to be undone in the opposite order they have been applied before~\cite{blocked-clause-undo}.

%\vspace{-2.5ex}
\paragraph{Variable Elimination} (VE)~\cite{satelite,niver-preprocessor} is a technique to remove variables from the formula. Removing a variable is done by resolving the according clauses in which the variable occurs. Given two sets of clauses: $C_x$ with the positive variable $x$ and $C_{\overline{x}}$ with negative $x$. Let $G$ be the union of these two sets $G \equiv C_x \cup C_{\overline{x}}$. Resolving these two sets on variable $x$ results in a new set of clauses $G'$ where tautologies are not included. It is shown in~\cite{satelite} that $G$ can be replaced by $G'$ without changing the satisfiability of the formula. If a model is needed for the original formula, then the partial model can be extended using the original clauses $F$ to assign variable $x$. Usually, applying VE to a variable results in a larger number of clauses. In state-of-the-art preprocessors VE is only applied to a variable if the number of clauses does not increase. The resulting formula depends on the order of the eliminated variables. \textit{Pure literal elimination} is a special case of VE, because the number of resolvents is zero.

%\vspace{-2.5ex}
\paragraph{Blocked Clause Elimination} (BCE)~\cite{blocked-clause-elimination} removes redundant \textit{blocked clauses}. A clause $C$ is blocked if it contains a blocking literal $l$. A literal $l$ is a blocking literal, if $\overline{l}$ is part of $C$, and for each clause $C' \in F$ with $\overline{l} \in C'$ the resolvent $C \otimes_l C'$ is a tautology~\cite{clause-eliminations,blocked-clause-elimination}.
Removing a blocked clause from $F$ changes the satisfying assignments~\cite{clause-eliminations}.
Since BCE is confluent, the order of the removals does not change the result~\cite{blocked-clause-elimination}.

%\vspace{-2.5ex}
\paragraph{Equivalence Elimination} (EE)~\cite{unhiding} removes a literal $l$ if it is equivalent to another literal $l'$. Only one literal per equivalence class is kept. Equivalent literals can be found by finding strongly connected components in the binary implication graph (BIG). The BIG represents all implications in the formula by directed edges $l \to l'$ between literals that occur in a clause [ $\overline{l}, l'$ ]. If a cycle $a \to b \to c \to a$ is found, there is also a cycle $\overline{a} \to \overline{b} \to \overline{c} \to \overline{a}$ and therefore $a \equiv b \equiv c$ can be shown and applied to $F$ by replacing $b$, and $c$ by $a$. Finally, double literal occurrences and tautologies are removed.

Let $F$ be $\langle [1, \neg 2]_1, [ \neg 1, 2]_2, [1, 2, 3]_3, [ \neg 1, \neg 3]_4, [ \neg 3, 4]_5, [ \neg 1, \neg 4 ]_6 \rangle$. The index $i$ of a clause $C_i$ gives the position of the clause in the formula. The order to apply techniques is EE, VE and finally BCE. EE will find $1 \equiv 2$ based on the clauses $C_1$ and $C_2$. Thus, it replaces each occurrence of $2$ with $1$, since $1$ is the smaller variable. This step alters $C_3$ to $C_7 = [1, 3]$. Now VE on variable $3$ detects that there are 3 clauses in which 3 occurs. The single resolvent that can be build is $C_{7\otimes5} = [1, 4]$. Finally, BCE removes the two clauses, because all literals of each clause are blocking literals. Since the resulting formula is empty, it is satisfied by any interpretation. It can be clearly seen, that the original formula cannot be satisfied by any interpretation.

%\vspace{-1ex}
\subsection{Equivalence-Preserving Techniques}

Equivalence-preserving techniques can be applied in any order, because the preprocessed formula is equivalent to the original one. By combining the following techniques with satisfiability-preserving techniques the order of the applied techniques has to be stored, to be able to undo all changes correctly.

%\vspace{-2.5ex}
\paragraph{Hidden Tautology Elimination} (HTE)~\cite{clause-eliminations} is based on the clause extension \textit{hidden literal addition} (HLA). After the clause $C$ is extended by HLA, $C$ is removed if it is tautology. The HLA of a clause $C$ with respect to a formula $F$ is computed as follows: Let $l$ be a literal of $C$ and $[l', l] \in F \setminus \{C\}$. If such a literal $l'$ can be found, $C$ is extended by $C := C \cup \overline{l'}$. This extension is applied until fix point. HLA can be regarded as the opposite operation of self subsuming resolution. The algorithm is linear time in the number of variables~\cite{clause-eliminations}. An example for HTE is the formula $F$ = $\langle [1, 3], [-2, 3], [1,2] \rangle$. Extending the clause $C_1$ stepwise can look as follows: $C_1 = [1, 3, \neg 2]$ with $C_3$. Next, $C_1 = [1, 3, \neg 2, 2]$ with $C_2$, so that it becomes tautology and can be removed.

%\vspace{-2.5ex}
\paragraph{Probing}~\cite{probing} is a technique to simplify the formula by propagating variables in both polarities $l$ and $\overline{l}$ separately and comparing their implications or by propagating all literals of a clause $C = [l_1, \dots, l_n]$, because it is known that in the two cases one of the candidates has to be satisfied.

Probing a single variable can find a conflict and thus finds a new unit. The following example illustrates the other cases:

\begin{center}
	\begin{tabular}{ r @{ } c @{ } l }
		BCP($1$)     & $\Rightarrow$ & 2, 3, 4, $\neg$5, $\neg$7 \\
		BCP($\overline{1}$) & $\Rightarrow$ & 2, $\neg$4, 6, 7 \\
	\end{tabular} 
\end{center}

To create a complete assignment, variable \textit{1} has to be assigned and both possible assignments imply $2$, so that $2$ can be set to \textit{true} immediately. Furthermore, the equivalences $4 \equiv 1$ and $\overline{7} \equiv 1$ can be found. These equivalences can also be eliminated. Probing all literals of a clause can find only new units.

%\vspace{-2.5ex}
\paragraph{Vivification} (also called \textit{Asymmetric Branching})~\cite{vivifying-propositional} reduces the length of a clause by propagating the negated literals of a clause $C = [l_1, \dots, l_n]$ iteratively until one of the following three cases occurs:
\begin{enumerate}
 \item BCP($\{\overline{l_1}, \dots, \overline{l_i}\}$) results in an empty clause for $i<n$.
 \item BCP($\{\overline{l_1}, \dots, \overline{l_i}\}$) implies another literal $l_j$ of the $C$ with $i < j < n$
 \item BCP($\{\overline{l_1}, \dots, \overline{l_i}\}$) implies another negated literal $\overline{l_j}$ of the $C$ with $i < j \leq n$
\end{enumerate}

In the first case, the unsatisfying partial assignment is disallowed by adding a clause $C' = [l_1, \dots, l_i]$. The clause $C'$  subsumes $C$. The implication $\overline{l_1} \land \dots \land \overline{l_i} \to l_j$ in the second case results in the clause $C' = [l_1, \dots, l_i,l_j]$ that also subsumes $C$. Formulating the third case into a clause $C' = [l_1, \dots, l_i,\overline{l_j}]$ subsumes $C$ by applying self subsumption to $C'' = C \otimes_{l_j} C' = [l_1, \dots, l_{j-1}, l_{j+1}, \dots, l_n]$.

%\vspace{-2.5ex}
\paragraph{Extended Resolution} (ER)~\cite{extendedResolution} introduces  a new variables $v$ to a formula that is equivalent to a disjunction of literals $v \equiv l \lor l'$. All clauses in $F$ are updated by removing the pair and adding the new variable instead. It has been shown, that ER is good for shrinking the proof size for unsatisfiable formulas. Applying ER during search as in~\cite{extendedResolution} resulted in a lower performance of \RISS, so that this technique has been put into the preprocessor and replaces the most frequent literal pairs. Still, no deep performance analysis has been done on this technique in the preprocessor, but it seems to boost the performance on unsatisfiable instances.

\vspace{-1.5ex}
\section{Coprocessor} \label{coprocessor}
The preprocessor of \RISS, \textit{Coprocessor}, implements all the techniques presented in Sect.~\ref{techniques} and introduces many algorithm parameters. A description of these parameters can be found in the help of \textit{Coprocessor}\footnote{The source code can be found at \url{www.ki.inf.tu-dresden.de/~norbert}.}. The techniques are executed in a loop on $F$, so that for example the result of HTE can be processed with VE and afterwards HTE tries to eliminate clauses again.

It is possible to maintain a blacklist and a white-list of variables. Variables on the white-list are tabooed for any non-model-preserving techniques so that their semantic is the same in $F'$. Variables on the blacklist are always removed by VE.

Furthermore, the resulting formula can be compressed. If variables are removed or are already assigned a value, the variables of the reduct of $F'$ are usually not dense any more. Giving the reduct to another solver increases its memory usage unnecessarily. To overcome this weakness, a compressor has been built into \textit{Coprocessor} that fills these gaps with variables that still occur in $F'$ and stores the already assigned variables for postprocessing the model. The compression cannot be combined with the white-list.

Another transformation that can be applied by the presented preprocessor is the conversion from encoded CSP domains from the direct encoding to the regular encoding as described in~\cite{CSPSAT2011}.

%\vspace{-1ex}
\subsection{The Map File Format}

A map file is used to store the preprocessing information that is necessary to postprocess a model of $F'$ such that it becomes a model for $F$ again. The map file and the model for $F'$ can be used to restore the model for $F$ by giving this information to \textit{Coprocessor}. The following information has to be stored to be able to do so:
\begin{center}
  \begin{tabular}{| l | l |}
  \hline
  \hfill \textbf{Once} \hfill ~& \hfill \textbf{Per elimination step} \hfill ~\\
  \hline
  Compression Table   & Variable Elimination \\
  \hline
  Equivalence Classes\hspace{1em} & Blocked Clause Elimination\\
  \hline
  ~ & Equivalence Elimination Step\hspace{1em} \\
  \hline
  \end{tabular}
\end{center}

The map file is divided into two parts. An partial example file for illustration is given in Fig.~\ref{map-example}. The format is described based on this example file. Each occurring case is also covered in the description. The first line has to state ``original variables" (line 1). This number is specified in the next line (line 2). Next, the compression information is given by beginning with either ``compress table" \mbox{(line 3)}, if there is a table, or ``no table", if there is no compression. Afterwards, the tables are given where each starts with a line \mbox{``table $k$ $v$"} and $k$ represents the number of the table and $v$ is the number of variables before the applied compression \mbox{(line 4)}. The next line gives the com-

\begin{wrapfigure}{l}{15em}
\vspace{-1.1em}
  \begin{center}
    \begin{verbatim}
 1:original variables
 2:30867
 3:compress tables
 4:table 0 30867
 5:1 2 3 5 6 7 9 10 11 ... 0
 6:units 0
 7:-31 32 ... -30666 -30822 0
 8:end table
 9:ee table
10:1 -19 0
11:2 -20 0
12:...
13:postprocess stack
14:ee
15:bce 523
16:-81 523 -6716 0
17:bce 10623
18:-10429 10623 -30296 0
19:...
20:ve 812 1
21:-812 -74 0
22:ve 6587 4
23:6587 6615 0
24:-79 6587 0
25:...
\end{verbatim}
\vspace{-1em}
  \end{center}
  \caption{Example map file\label{map-example}}
%\vspace{-4em}
\end{wrapfigure}

\noindent
pression by simply giving a mapping that depends on the order: the $i^{th}$ number in the line is the variable that is represented by variable $i$ in the compressed formula \mbox{(line 5)}. The line is closed by a $0$, so that a standard clause parser can be used. The next line introduces the assignments in the original formula by saying ``units $k$" \mbox{(line 6)}. The following line lists all the literals that have been assigned \textit{true} in the original formula and is also terminated by $0$ \mbox{(line 7)}. The compression is completed with a line stating \mbox{``end table"} \mbox{(line 8)}.
At the moment, only a single compression is supported, and thus, $k$ is always 0. Since there is only a single compression, it is applied after applying all other techniques and therefore the following details are given with respect to the decompressed preprocessed formula $F'$. The next static information is the literals of the EE classes.
They are introduced by a line \mbox{``ee table"} \mbox{(line 9)}. The following lines represent the classes where the first element is the representative of the class that is in $F'$\mbox{(line 10-12)}. Each class is ordered ascending, so that the EE information can be stored as a tree and the first element is the smallest one. Again, each class is terminated by a 0. Finally, the postprocess stack is given and preluded with a line ``postprocess stack" \mbox{(line 13)}. Afterwards the eliminations of BCE and VE are stored in the order they have been performed. BCE is prefaced with a line \mbox{``bce $l$"} where $l$ is the blocking literal \mbox{(line 15,17)}. The next line gives the according blocked clause \mbox{(line 16,18)}. For VE the first line is \mbox{``ve $v$ $n$"} where $v$ is the eliminated variable and $n$ is the number of clauses that have been replaced \mbox{(line 20,22)}. The following $n$ lines give the according clauses \mbox{(line 21,23-26)}. Finally, for EE it is only stated that EE has been applied by writing a line ``ee", because postprocessing EE depends also on the variables that are present at the moment \mbox{(line 14)}. Some of the variables might already be removed at the point EE has been run, so that it is mandatory to store this information.

\subsection{Preprocessor Comparison}

\begin{figure}[t!]
\centering
\includegraphics[page=1,keepaspectratio=true,totalheight=7.5cm]{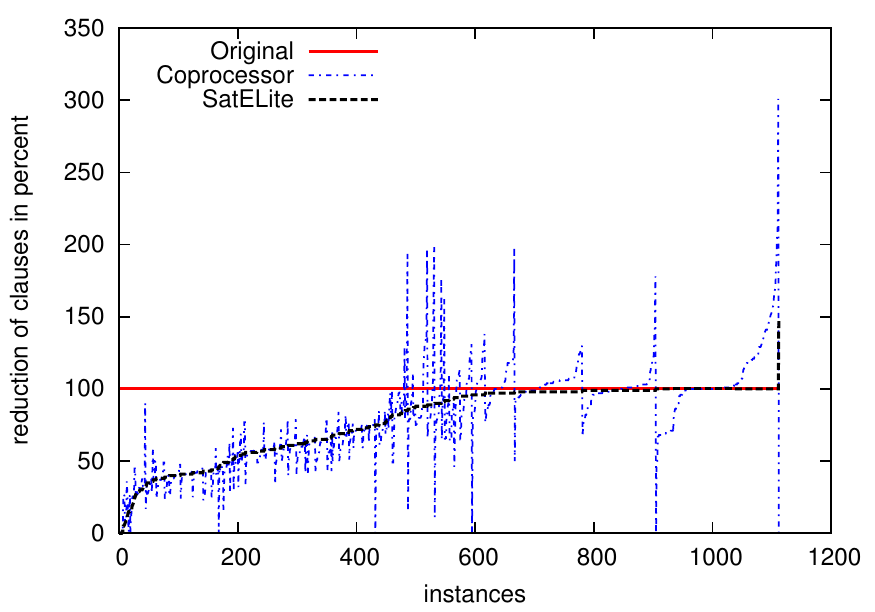}
\caption{Relative reduction of SatELite and Coprocessor\label{preprocessor-comparison}}
%\vspace{-4ex}
\end{figure}

A comparison of the formula reductions of \textit{Coprocessor} and the current standard preprocessor SatELite is given in Fig.~\ref{preprocessor-comparison} and has been performed on $1155$ industrial and crafted instances from recent SAT Competitions and SAT Races\footnote{For more details visit  \url{http://www.ki.inf.tu-dresden.de/~norbert/paperdata/WLP2011.html}.}. The relative reduction of the clauses by \textit{Coprocessor} and SatELite is presented. Due to ER, \textit{Coprocessor} can increase the number of clauses, whereby the average length is still reduced. \textit{Coprocessor} is also able to reduce the number of clauses more than SatELite. The instances are ordered by the reduction of SatELite so that the plot for \textit{Coprocessor} produces peaks. 

\begin{figure}[h!]
\centering
 \includegraphics[page=1,keepaspectratio=true,totalheight=7.5cm]{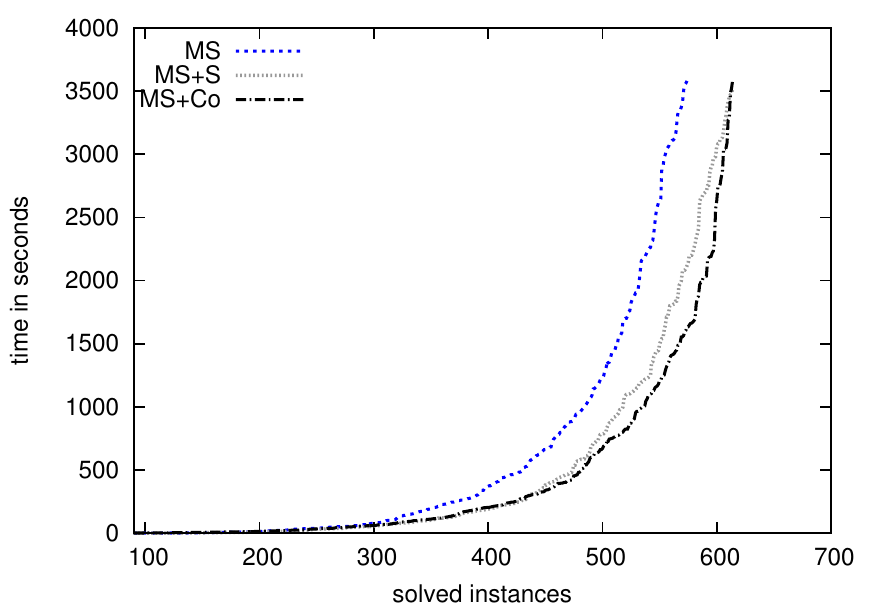} 
\caption{Runtime comparison of MiniSAT combined with Coprocessor and SatELite\label{preprocessor-runtime}}
%\vspace{-4ex}
\end{figure}

Since SatELite~\cite{satelite} and \textit{MiniSAT}~\cite{minisat22} have been developed by the same authors, the run times of \textit{MiniSAT} with the two preprocessors are compared in Fig.~\ref{preprocessor-runtime}. Comparing these run times of \textit{MiniSAT} (MS) combined with the preprocessors, it can be clearly seen that by using a preprocessor the performance of the solver is much higher. Furthermore, the combination with \textit{Coprocessor} (MS+Co) solves more instances than \textit{SatELite} (MS+S) for most of the timeouts.

\newpage

%\vspace{-1.5ex}
\section{Conclusion and Future Work} \label{conclusion}
\ifx \SHORT\relax
\begin{itemize}	
	\item parallel important, hopefully better, scalable
\end{itemize}
\fi

This work introduces the SAT preprocessor \textit{Coprocessor} that implements almost all known preprocessing techniques and some additional features. Experiments showed that the default \textit{Coprocessor} performs better than \textit{SatELite} when combined with \textit{MiniSAT 2.2}. For suiting its techniques better to applications, \textit{Coprocessor} provides many parameters that can be optimized for special use cases. Additionally, a map file format is presented that is used to store the preprocessing information. This file can be used to re-construct the model for the original formula if the model for the preprocessed formula is given.

Future development of this preprocessor includes adding the latest techniques such as HLE and HLA~\cite{clause-eliminations,unhiding} and to parallelize it to be able to use multi-core architectures. Furthermore, the execution order of the techniques will be relaxed, so that any order can be applied to the input formula.

\emph{Acknowledgment}\ \ 
The author would like to thank Marijn Heule for providing an implementation of HTE and discussing the dependencies of the algorithms.
%\vspace{-0.7ex}

%\begin{footnotesize}
%\vspace{-2ex}
%\bibliographystyle{plain}
\bibliographystyle{abbrv}
\bibliography{ref/refs,ref/applications}
%\end{footnotesize}

\newpage

%\appendix
%
%\section{Problem Instances Used in the Measurement}\label{benchmark-table}
%\input{data/benchmark-table.tex}

\end{document}